\documentclass[final]{article}

\usepackage[nonatbib]{neurips_2023}

\usepackage[utf8]{inputenc} 
\usepackage[T1]{fontenc}    
\usepackage{hyperref}       
\usepackage{url}            
\usepackage{booktabs}       
\usepackage{amsmath,amsfonts}
\usepackage{nicefrac}       
\usepackage{microtype}      
\usepackage{xcolor}         
\usepackage{natbib}
\usepackage{graphicx}
\usepackage{algorithm}
\usepackage{algpseudocodex}
\usepackage{orcidlink}

\DeclareMathOperator{\ang}{{\rm \textup{\AA}}}

\title{Multiscale Feature Attribution for Outliers}

\author{%
  Jeff Shen\,\orcidlink{0000-0001-6662-7306} \\
  Department of Astrophysical Sciences \\
  Princeton University\\
  \texttt{shenjeff@princeton.edu} \\
  \And
  Peter Melchior\,\orcidlink{0000-0002-8873-5065} \\
  Department of Astrophysical Sciences \\
  Center for Statistics \& Machine Learning\\
  Princeton University\\
  \texttt{peter.melchior@princeton.edu} \\
}

\begin{document}

\maketitle

\begin{abstract}
Machine learning techniques can automatically identify outliers in massive datasets, much faster and more reproducible than human inspection ever could. But finding such outliers immediately leads to the question: which features render this input anomalous?
We propose a new feature attribution method, Inverse Multiscale Occlusion, that is specifically designed for outliers, for which we have little knowledge of the type of features we want to identify and expect that the model performance is questionable because anomalous test data likely exceed the limits of the training data.
We demonstrate our method on outliers detected in galaxy spectra from the Dark Energy Survey Instrument and find its results to be much more interpretable than alternative attribution approaches.
\end{abstract}

\section{Introduction}
\label{sec:intro}

Machine learning (ML) is becoming increasingly necessary in astronomy to deal with the massive datasets that are already available or will become available in the near future.
But the adoption of novel tools does change the scientific motivation, namely to \emph{understand} the phenomena in the data. It is therefore important to investigate not just if a ML method works, or how well it works, but also why it produces the reported results.
For complex methods, such as deep neural networks, explainability is an open challenge, which suggest a rephrasing of the question: what aspects of the data led to the reported results? Several feature attribution approaches have been developed to answer this question and yield insights not just into the data, but also into the inner workings of the ML method.

We investigate feature attribution methods for the task of outlier detection. ML methods are very capable of finding anomalous data in large data sets, but every detection immediately begs the question: what made that input an outlier?
We present a new feature attribution method and compare it to existing alternatives on outliers identified by \citet{Liang2023a} in spectra of the Early Data Release \citep[EDR; ][]{DESICollaboration2023} of the Dark Energy Survey Instrument \citep[DESI; ][]{DESICollaboration2016} Bright Galaxy Survey \citep[BGS; ][]{Hahn2023}.

The outlier detection method is based on the autoencoder model by \citet{Melchior2022}. It encodes spectra into a low-dimensional latent space and then uses a normalizing flow in the latent space to determine objects that have low probability relative to the rest of the data \citep[see][for details]{Liang2023}.
We now want to identify---without human intervention---the features in a spectrum that most contribute to the ML model (here the combination of autoencoder and normalizing flow) identifying that spectrum as an outlier.

\section{Feature Attribution Methods}
\label{sec:attribution}

Numerous methods have been proposed for feature attribution in machine learning. 
We discuss two main classes, beginning with gradient-based attribution methods. 
The most direct one, {\bf Instantaneous Gradients}, considers only the gradient of the output $f(\mathbf{x})\in \mathbb{R}^L$, in our case the likelihood $\log p(\mathbf{x})$ from a normalizing flow.
An early example of this approach is the saliency maps of \citet{Simonyan2014}. 
While extremely fast and simple to implement, instantaneous gradients may saturate when a change in input does not lead to a change in output \citep{Shrikumar2019}.
\citet{Sundararajan2017} proposed {\bf Integrated Gradients (IG)} to solve this saturation issue.
The idea is to integrate the gradients of the output with respect to the input along a straight line (i.e., interpolations) from a baseline input to the observed input.
Doing so requires numerical integration over the interpolations between the baseline and the observed input, which make this is a slower method.
The choice of the baseline is not trivial \citep[see][for an extensive discussion]{Sturmfels2020}. 
For our particular application, the common baseline of a zero vector will turn out to be inappropriate (see Section \ref{sec:method}). 
{\bf Expected gradients (EG)} mitigate the choice of baseline by averaging over many different baselines drawn randomly from the underlying dataset \citep{Erion2020}.

The second class of methods perturb the inputs.
{\bf Feature ablation (FA)} replaces a single input feature with its corresponding value from some baseline, forming a modified spectrum $\mathbf{\hat x}$, and calculates the difference between the two versions, $\mathrm{FA} = f(\mathbf{x}) - f(\mathbf{\hat x})$, repeating the same procedure for all features.
If changing the feature value has a large impact on the output, then that feature is important. 
However, input features are often correlated.
In our case, we know that the majority of variation in spectra can be modelled by very few variables, a property that is exploited by the autoencoder compression that the outlier detection model of \citet{Liang2023a} is based on.
Furthermore, many interesting features are wider than a single pixel (e.g., broad lines from quasars).
The ablation of a single feature may therefore miss many important features.
{\bf Occlusion} is an attribution method similar to ablation; it has been used in computer vision tasks \citep[e.g., ][]{Zeiler2013} and involves ablating a contiguous block of size $W$ with a baseline rather than a single feature.

\section{Feature Attribution for Outliers}
\label{sec:method}

We propose {\bf Inverse Multiscale Occlusion (IMO)} for feature attribution in detected outliers, a modification of standard occlusion with several desirable features. Here we describe the motivations for the development of this algorithm and provide the pseudocode in Algorithm \ref{alg:imo}.

\begin{figure}[!tbp]
  \centering
  \includegraphics[width=\textwidth]{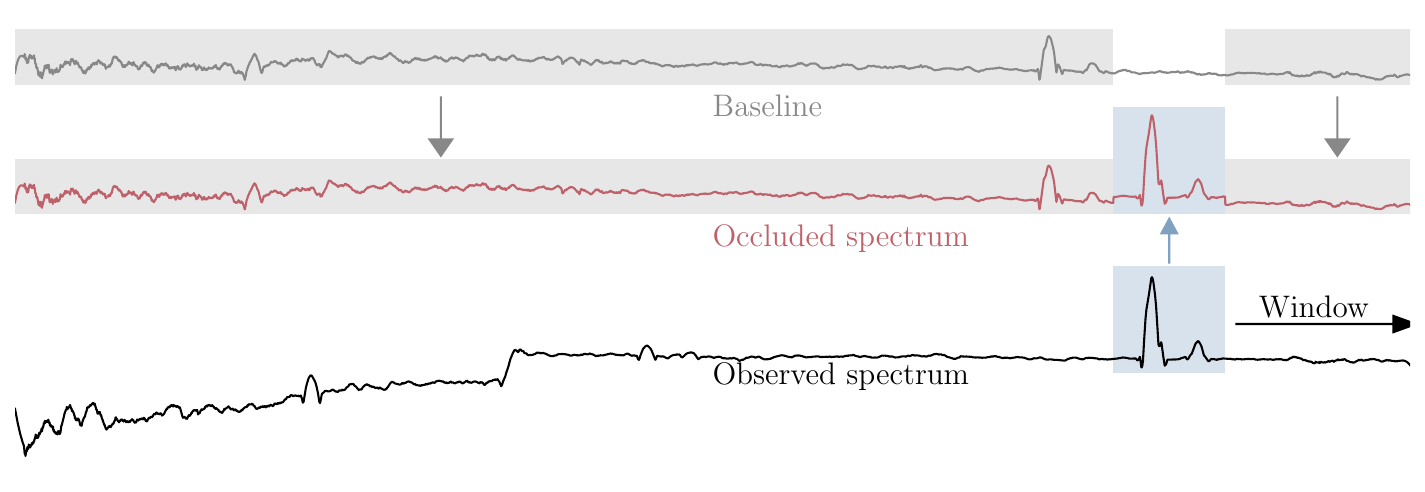}
  \caption{In inverse occlusion, a segment of the baseline is replaced with a segment of the test spectrum to create an occluded spectrum, which is then used to calculate the network output. This approach is more robust for outliers because the bulk of the occluded spectrum is not anomalous.}
  \label{fig:inverse-occlusion}
\end{figure}

Outliers have by construction low probability.
Modifying an already anomalous input further by occlusion with a baseline spectrum will likely lead to erratic behavior because the modified feature vector $\mathbf{\hat x}$ is almost guaranteed to be outside of the training data.
We therefore perform ``inverse occlusion'', which replaces a segment of length $W$, centered at position $i$, of the baseline spectrum $\mathbf{x}_b$ with the corresponding segment of the input spectrum to form the modified spectrum $\mathbf{\hat x}_{\overline{W_i}}$ (see Figure \ref{fig:inverse-occlusion}), and use the probability of the baseline, rather than the input spectrum, as the reference.
For a given position, window size, and baseline $b$, the attribution metric is thus $\mathrm{IMO}_{W_i,b}=f(\mathbf{x}_b) - f(\mathbf{\hat x}_{\overline{W_i}})$.
By sliding this window across the entire input, we get a vector $\mathrm{IMO}_{W,b}\in\mathbb{R}^L$. 
Larger values of $\mathrm{IMO}_W$ indicate a higher probability for the original baseline not containing a particular segment from the input spectrum, i.e., this segment is unusual.

Anomalous features may be broad (spread over multiple pixels) or narrow. 
In the case of galaxy spectra, contamination by another source is a broad feature because the entire spectrum will look different, but outflows causing asymmetric emission lines is a much more localized feature.
We want our method to identify the relevant features in both cases without having to specify a priori what we are looking for.
Taking inspiration from wavelet filtering, we perform occlusion with variable window sizes, i.e., we get $\mathrm{IMO}_W$ for each choice of window size $W$.
Smaller window sizes respond to local features, while larger windows sizes are sensitive to broad or even global features.
We compute attributions for all window sizes, so that users can decide what features are relevant in each case.

This leaves the question as to how to choose the baseline, which serves as a counterfactual to determine how the model would have responded to an input if it did not have that feature.
We adopt the use of multiple baselines, randomly drawn from the underlying dataset, from the EG method, but we make a domain-specific adjustment for galaxy spectra.
We need to ensure that the baselines are consistent with the input spectrum even if they come from galaxies at different redshifts, i.e., where spectral features are stretched by a different amount from the input spectrum. To remove this source of discrepancy, we redshift the baseline spectra to the same redshift as the input spectrum.
In other words, we are finding anomalous features for objects at a particular redshift because we are comparing to objects at the same redshift.

The use of an ensemble of baselines $\mathcal{B}$ also enables us to form a single feature attribution result as the minimum-variance combination across all window sizes:
\begin{equation*}
    \begin{split}
    \mathrm{IMO_W} &= \mathbb{E}_{b\sim\mathcal{B}}\left[\mathrm{IMO}_{W,b}\right],\ \
   \sigma_W^2 = \mathrm{var}_{b\sim\mathcal{B}}\left[\mathrm{IMO}_{W,b}\right]\\
   \mathrm{IMO} &= \sum_W \mathrm{IMO}_W \odot \sigma_W^{-2} \Big/ \sum_W \sigma_W^{-2}, 
    \end{split}
\end{equation*}
with all operations computed element-wise in $\mathbb{R}^L$. 
This attribution method is emphasizes, at every location $i$, the window size for which the most significant discrepancies are found in the ensemble.

\section{Application to DESI Outliers}
\label{sec:desi}

\begin{figure}[!tbp]
  \centering
  \includegraphics[width=\textwidth]{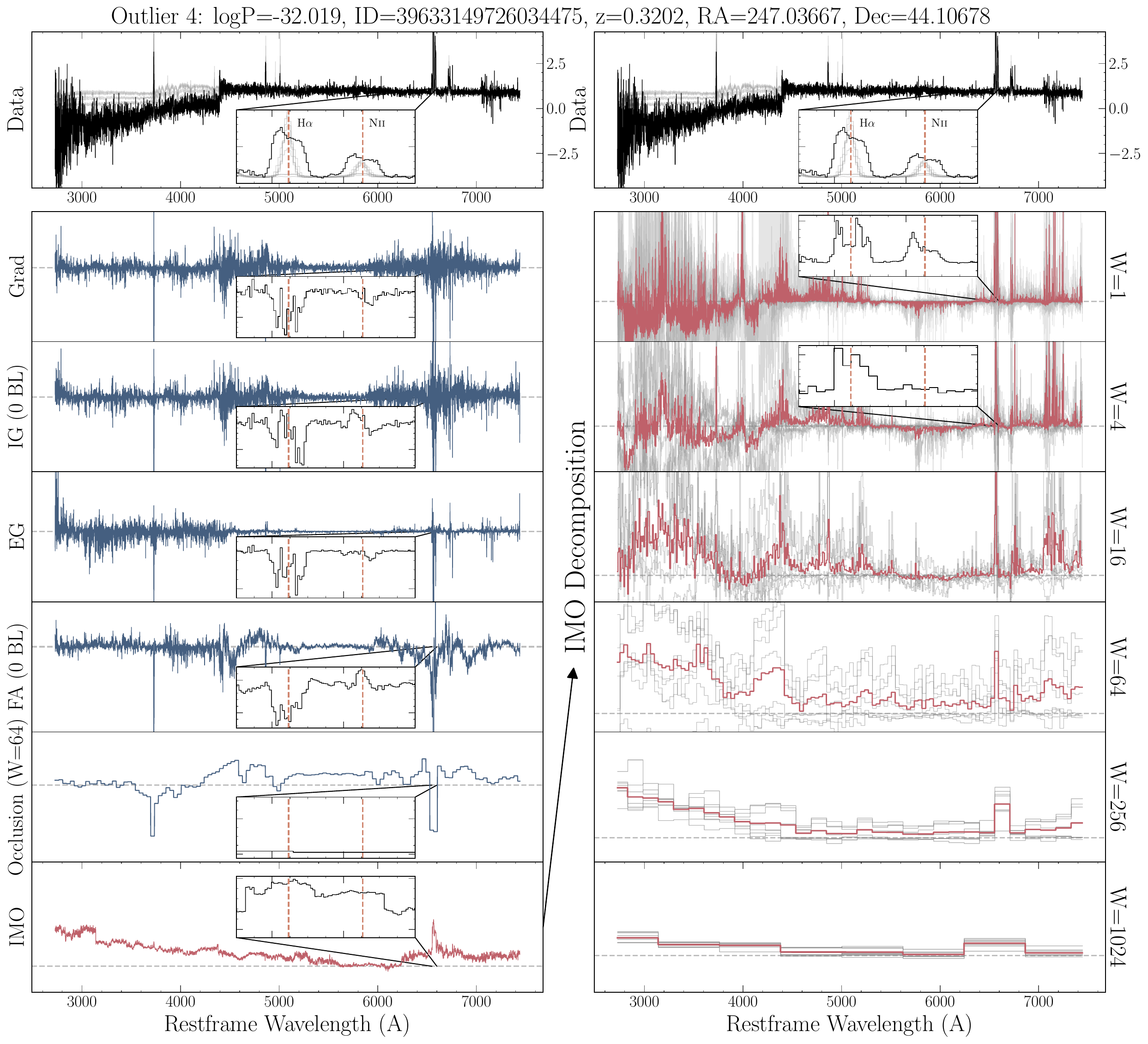}
  \caption{Comparison of attribution methods for spectroscopic outlier 4 from \citep{Liang2023a}, which exhibits a large-scale calibration problem and double-peaked emission lines.
  The observed spectrum is shown in the top panel, with several baselines in light gray.
  Existing feature attributions are shown in blue in the left column; the last panel shows the our combined Inverse Multiscale Occlusion (IMO) in red.
  The right column shows the IMO attributions separately for each window size, with individual baseline results in gray. The insets zoom in on a region around the H$\alpha$ line.} 
  \label{fig:feature-attribution-4}
\end{figure}

We apply the feature attribution methods we discussed above to the DESI spectrum of outlier 4, as listed by \citet{Liang2023a}.
As shown in the top panel of Figure \ref{fig:feature-attribution-4}, the spectrum shows a calibration problem in the B arm of spectrograph, causing a jump and negative flux values on the blue side of the spectrum.
The same spectrum also shows double-peaked emission lines, indicating gas that moves with different relative velocities. 
This is a localized anomalous feature, particularly visible in the emission of H$\alpha$ around $6563\ang$ (insets in Figure \ref{fig:feature-attribution-4}).
With large- and small-scale anomalous features, outlier 4 is a good test case for attribution methods.

Gradient-based methods show no significant attributions in the blue part of the spectrum, largely missing the obvious calibration problem.
FA shows a significant bump at the transition between the anomalous B arm and the correct R arm of the spectrograph, but becomes flat bluewards of $4500\ang$, missing the progressive deviation towards the blue end.
IMO detects this large-scale anomaly.
Looking at right panel of Figure \ref{fig:feature-attribution-4}, we can see that the sharp jump is picked up by moderate window sizes of $W=16$ and $W=64$, while the progressive drop towards the blue end shows up when using larger window sizes.
With a single window size, e.g., for standard occlusion with $W=64$, the calibration mismatch does not show up as a prominent feature.

The emission line shapes, especially for H$\alpha$ and \textsc{Nii}, are detected as anomalous in almost all methods.
The gradient-based methods show negative attributions on either side of the H$\alpha$ line, indicating that suppressing the data there would make the spectrum appear less anomalous.
By averaging over multiple baselines, EG produces the least noisy attributions.
FA also detects the double-peaked line shapes, with negative values indicating a region where the test spectrum has lower probability than the baseline.
On the other hand, standard occlusion with $W=64$ recognizes the line region itself as somewhat anomalous, but fails to resolve any details because the window size is too large compared to the size of the features.
Finally, for IMO the H$\alpha$ line region produces the highest outlier attribution.
In the combined statistic, in the bottom left panel of Figure \ref{fig:feature-attribution-4}, the line appears somewhat broadened, a consequence of having significant attributions over a range of window sizes.
Decomposing the outlier attribution into different window sizes reveals a very clear double-peak anomaly for $W=1$ and $W=4$, with peaks on either side of the nominal line center, demonstrating the utility of reporting all $\mathrm{IMO}_W$ separately.
Finally, IMO with $W=1$ localizes the major anomalies as distinct peaks much more evidently as FA, despite both being based on single-pixel occlusion.
This result is a clear demonstration of the benefit of the inverted occlusion method. 
In IMO, the baseline is (typically) not anomalous, and the occluded vector is thus likely in the region of good model performance, so that anomalous features produce more meaningful attributions.

\section{Conclusion}

Our new attribution method, Inverse Multiscale Occlusion, is able to identify anomalous features---both large-scale and small-scale---for known outliers.
By applying IMO to a spectroscopic outlier in DESI BGS, we demonstrate that it peaks at a large calibration offset and at double-peaked emission lines, reproducing the findings from the visual inspection performed by \citet{Liang2023a}.
We show that IMO outperforms gradient-based methods, which often capture only highly localized anomalies and are harder to interpret, and is also more flexible and better suited to outliers than other perturbative methods.
IMO produces a meaningful outlier attribution summary as well as scale-dependent attributions, which enables the interpretation of anomalies with little human supervision. 

\begin{ack}

\end{ack}

\section{Supplementary Material}

We list the pseudo-code of our feature attribution method IMO below:
\algnewcommand{\Inputs}[1]{%
  \State \textbf{Inputs:}
  \Statex \hspace*{\algorithmicindent}\parbox[t]{.8\linewidth}{\raggedright #1}
}

\begin{algorithm}
  \caption{Inverse multiscale occlusion (IMO)}\label{alg:imo}
  \begin{algorithmic}[1]
    \State \textbf{Inputs:}
    \Statex \hspace*{\algorithmicindent} Input spectrum $\mathbf{x} \in \mathbb{R}^N$
    \Statex \hspace*{\algorithmicindent} Input redshift $z \in \mathbb{R}$
    \Statex \hspace*{\algorithmicindent} Encoder model $f: \mathbb{R}^N \to \mathbb{R}^S$
    \Statex \hspace*{\algorithmicindent} Decoder model $g: (\mathbb{R}^S, \mathbb{R}) \to \mathbb{R}^N$
    \Statex \hspace*{\algorithmicindent} Normalizing flow log-probability function $h: \mathbb{R}^S \to \mathbb{R}$
    \Statex \hspace*{\algorithmicindent} Spectrum dataset $\mathcal{D}$
    \Statex \hspace*{\algorithmicindent} Number of baseline samples $M$
    \Statex \hspace*{\algorithmicindent} Window sizes $W = {W_1, \ldots, W_K}$
    \Function{InverseMultiscaleOcclusion}{$\mathbf{x}, z, \mathcal{D}, M, W$}
    \State \textbf{Initialize:} 
    \Statex \hspace*{\algorithmicindent} $\mathbf{A} \gets \texttt{zeros}(K, N)$, $\mathbf{p} \gets \texttt{zeros}(N)$, $\mathbf{R} \gets \texttt{zeros}(M, N)$
    \For{$j = 1, \ldots, M$}
      \Comment{Loop over all baselines}
      \State $\mathbf{B} \sim \mathcal{D}$
      \Comment{Sample a baseline spectrum from the dataset}
      \State $\mathbf{R_j} \gets g(f(\mathbf{B}), z)$
      \Comment{Reconstruct the baseline spectrum at the input redshift}
      \State $p_j \gets h(f(\mathbf{R_j}))$
      \Comment{\parbox[t]{.5\linewidth}{Encode and compute the log-probability of the redshifted baseline spectrum}}
    \EndFor
    \State \textbf{end for}
    \For{$k = 1, \ldots, K$}
      \Comment{Loop over all window sizes}
      \State $\mathbf{T} \gets \texttt{zeros}(M, N)$
      \Comment{Initialize temporary attribution array}
      \For{$j = 1, \ldots, M$}
        \Comment{Loop over all the baseline samples}
        \State $i \gets 0$
        \Comment{Initialize the counter index}
        \For{$c = 1, \ldots, \texttt{ceil}(N / W_k)$}
          \Comment{Loop over all chunks}
          \State $u \gets \min(i + W_k, N)$
          \Comment{Calculate the upper bound of the current chunk}
          \State $\mathbf{y} \gets \mathbf{R_j}$
          \Comment{Copy the baseline spectrum}
          \State $\mathbf{y}_{i:u} \gets \mathbf{x}_{i:u}$
          \Comment{\parbox[t]{.5\linewidth}{Replace the current chunk with the corresponding chunk from the input spectrum}}
          \State $l \gets h(f(\mathbf{y}))$
          \Comment{Compute the log-probability of the occluded spectrum}
          \State $T_{j, i:u} \gets (p_j - l) / (u - i)$
          \Comment{\parbox[t]{.3\linewidth}{Calculate the attribution for the current chunk}}
          \State $i \gets i + W_k$
          \Comment{Increment the counter index}
        \EndFor
        \State \textbf{end for}
      \EndFor
      \State \textbf{end for}
      \State $\mathbf{A_k} \gets \frac{1}{M} \sum_{j=1}^M \mathbf{T_{j}}$
      \Comment{Compute average attribution for each window size}
    \EndFor
    \State \textbf{end for}
    \State \Return $\mathbf{A}$
    \EndFunction
  \end{algorithmic}
\end{algorithm}

{
\small
\bibliography{sources-zotero, sources}
}

\end{document}